\documentclass[12pt]{article}
\usepackage{amsmath, amsfonts, amsthm, amssymb, epsfig}
\usepackage[english]{babel}
\usepackage[margin=1in]{geometry}
\usepackage{multicol}
\usepackage{background}
\usepackage{tabularx}
\usepackage[math]{blindtext}
\usepackage{float}
\restylefloat{table}
\usepackage{caption}
\usepackage{subcaption}

\SetBgScale{4}
\SetBgColor{gray}
\SetBgAngle{90}
\SetBgContents{arXiv: some other text goes here}
\SetBgPosition{-1.5,-10}
\begin{document}

{\centering

{\bfseries\Large Visual Objects Classification with Sliding Spatial Pyramid Matching\bigskip}

Hao Wooi Lim\textsuperscript{1} and Yong Haur Tay\textsuperscript{2} \\
\normalfont 
	\texttt{\textsuperscript{1}  haowooilim@1utar.my, \textsuperscript{2} tayyh@utar.edu.my}\\
	
 {\itshape
Computer Vision \&\ Intelligent Systems (CVIS) group, Faculty of Engineering and Science, University of Tunku Abdul Rahman\\
  }
}

\begin{abstract}
We present a method for visual object classification using only a single feature, transformed color SIFT \cite{paper15} with a variant of Spatial Pyramid Matching (SPM) that we called Sliding Spatial Pyramid Matching (SSPM), trained with an ensemble of linear regression (provided by LINEAR) to obtained state of the art result on Caltech-101 \cite{paper22} of 83.46\%. SSPM is a special version of SPM where instead of dividing an image into \(K\) number of regions, a subwindow of fixed size is slide around the image with a fixed step size. For each subwindow, a histogram of visual words is generated. To obtained the visual vocabulary, instead of performing K-means clustering \cite{paper26}, we randomly pick \(N\) exemplars from the training set and encode them with a soft non-linear mapping method from \cite{paper16}. We then trained 15 models, each with a different visual word size with linear regression. All 15 models are then averaged together to form a single strong model.
\bigskip

\end{abstract}


\section{Introduction}
Recently, there has been a lot of development in the area of Deep Learning
(DL). Not unlike traditional Multi-layer Neural Networks, DL sought to solve the difficult problems such as visual objects classification by utilizing many layers of neurons, which has been traditionally too computationally intensive to train, with the help of fast, many-cores Graphics Processing Unit (GPU) to help speedup training. One interesting outcome of the research in DL is such that, DL allows for learning of feature detectors, thus mostly eliminating the need for hand-engineered features. And sure enough, recent papers such as \cite{paper18} has shown that DL could even obtained state of the art results on various visual object data set such as MNIST \cite{paper19}, NIST SD 19 \cite{paper20}, CIFAR-10 \cite{paper21} and etc. However, noticed that most of the data set tested are usually consists of very small images. This is due to the fact that data set with huge dimensionality would pose an even bigger challenge for GPU, sometimes even requiring weeks of training, even with a cluster of GPUs. Moreover, not everyone has a huge cluster of GPUs at their disposal.

\begin{figure}[t]
\centering
\includegraphics[totalheight=0.8\textheight,width=11cm,height=7cm]{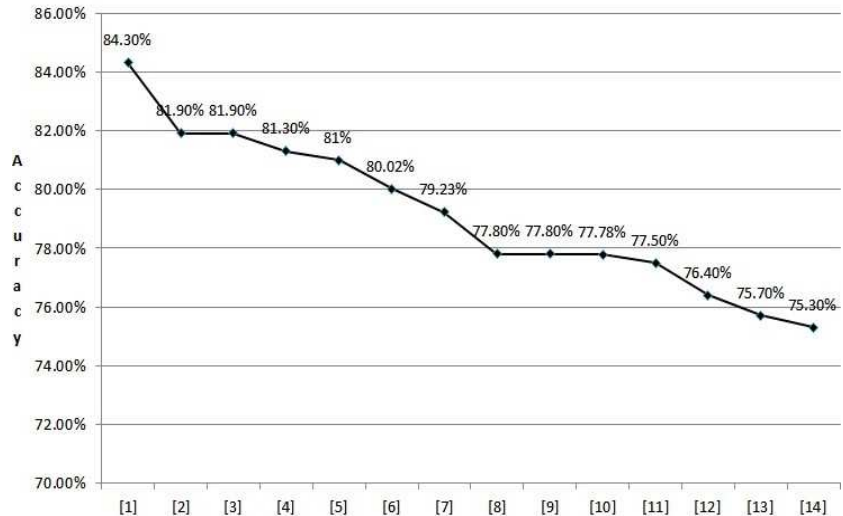}
\caption{Figure compares the accuracy on Caltech-101 data set, all trained on 30 samples.}
\label{fig:fig1}
\end{figure}

Looking at some of the best-performing methods (\cite{paper1}, \cite{paper4}, \cite{paper8}, \cite{paper9}), shown on Fig. \ref{fig:fig1} on Caltech-101 \cite{paper22}, one trend seems apparent: many different features are usually
employed to obtained the published results, most notably Scale-invariant feature transform (SIFT) \cite{paper23}, Pyramid histogram of orientated gradients (PHOG) \cite{paper9}, Self-similarity (SSIM) \cite{paper25}, Geometric Blur (GB) \cite{paper24} and etc. While it is easy to obtain better accuracy by employing many different features, it significantly increases the computation power required to train
them.

Some
algorithm like kernel-based Support Vector Machine (SVM) which requires 
\(O(nd)\) testing time, where \(n\) is the number of support vectors and
\(d\) is the dimensionality of the feature vector \cite{paper27}, it is just not feasible when the model is complex, resulting in large number
of support vectors or when the feature dimension is large. Unfortunately,
it is pretty easy to feed a lot of features for training, allowing the
training process to discover salient features useful for accurate classification.  Thus, in most cases, it is difficult to obtain good results with kernel-based
 SVM without also making training intractable.

In this paper, our goal is to lower the error bound siginificantly on challenging visual object data set such as Caltech-101 \cite{paper22} that has high enough number of categories and dimensionality, while also employing methods that are much more efficient to train than DL-based methods, relying on hand-engineered feature but refrained from combining many different kind of features so that training remains quick and efficient.

\section{Our method}
Here, we describe our method, along with the best parameters that we have found.

Our method begins with extraction of features. Following the method employed by \cite{paper15}, we use Harris-Laplace salient point detector to localize all the local keypoints in each image. For each keypoint, we use the Transformed Color SIFT as the descriptor.

To obtained a visual vocabulary, we randomly pick \(N\) exemplars from the training set instead of performing K-means clustering \cite{paper26}, which was the usual method employed by most paper.

To encode keypoint into feature vector, we encode them with a soft non-linear mapping method as described in \cite{paper16}.

Next, we use a special version of Spatial Pyramid Matching (SPM) where instead of dividing an image into \(K\) number of regions whereby for each region, a histogram of visual words is generated, a subwindow of fixed size is  slide around the image with fixed step size. Each subwindow then, is process the same as you would with a region in SPM. We call this Sliding Spatial Pyramid Matching (SSPM). Here, we set the width and height of the subwindow to be one-third of the original image's width and height respectively, while step size x and y to be the one-third of the subwindow's width and height respectively.

Since for each region we obtained a feature vector with size equal to the visual vocabulary size, we append each feature vector from each region into a single feature vector to be trained by linear regression.

To train the system, we used LIBLINEAR \cite{paper17} for this purpose (Due to the high dimensionality of our feature vector, kernel-based SVM is unfeasible). We choose L2-regularized logistic regression (primal) with probabilistic output.

Finally, to improve the accuracy, we trained multiple models and combined them by multiplying together the probabilistic output to obtained the final
probability distribution over class label.

\section{Data sets and Experimental Protocol}
We conduct our experiment on Caltech-101 \cite{paper22}. The data set consists of images from 101 object categories, and contains from 31 to 800 images per category. Most images are medium resolution, about 300 \(\times\) 300 pixels. The significance of this database is its large inter-class variability.
Following standard procedures, the Caltech-101 data is split into 30 training
images (chosen randomly) per category and 50 for testing. BACKROUND\_GOOGLE
category was not included in the experiment. The classification process is
repeated 4 times (changing the training and test sets) and the average performance score is reported.

\section{Results}
We study the influence of a variable visual vocabulary size (\(N=\{1000,900,...,100\}\)) or fixed visual vocabulary size (\(N=1000\)) on several models that was then averaged together to obtain a final result. Note that each model has a different visual vocabulary due to the fact that visual word is obtained with randomly chosen exemplars.
To speedup experiment, SIFT is used in this case.

\begin{table}[t]
\centering
\begin{tabular}{|c|c|c|}\hline
Number of models & Fixed visual vocabulary size & Variable visual vocabulary
size \\\hline
1 & 71.13\%\ & 70.92\%\ \\\hline
2 & 73.18\%\ & 72.95\%\ \\\hline
3 & 73.37\%\ & 72.70\%\ \\\hline
4 & 73.67\%\ & 72.97\%\ \\\hline
5 & 73.61\%\ & 73.69\%\ \\\hline
6 & 73.65\%\ & 73.57\%\ \\\hline
7 & 73.67\%\ & 73.54\%\ \\\hline
8 & 73.80\%\ & 73.71\%\ \\\hline
9 & 73.69\%\ & 73.48\%\ \\\hline
10 & 73.78\%\ & \textbf{73.84\%}\ \\\hline
\end{tabular}
\caption{Table compares the accuracy using variable visual vocabulary size vs fixed visual vocabulary size.}
\label{tab:tab1}
\end{table}

As shown in Tab. \ref{tab:tab1}, variable visual word seems to performed
slightly better. This makes sense, when combining several models together,
models that are different from each other tends to performed better after being combined. 

We then experiment with different step size and subwindow size. For method A, we set the width and height of the subwindow to be half of the original image's width and height respectively, while step size x and y to be the half of the subwindow's width and height respectively. For method B, we set the width and height of the subwindow to be one-third of the original image's width and height respectively, while step size x and y to be the one-third of the subwindow's width and height respectively. As one might exect, method B would result in features with higher dimension compared to method A, due to the fact that method B generates more regions.
To speedup experiment, SIFT is used in this case.

\begin{table}[t]
\centering
\begin{tabular}{|c|c|c|}\hline
Number of models & Method A & Method B \\\hline
1 & 70.92\%\ & 72.78\%\ \\\hline
2 & 72.95\%\ & 73.95\%\ \\\hline
3 & 72.70\%\ & 74.65\%\ \\\hline
4 & 72.97\%\ & 74.90\%\ \\\hline
5 & 73.69\%\ & 75.03\%\ \\\hline
6 & 73.57\%\ & 75.45\%\ \\\hline
7 & 73.54\%\ & 75.30\%\ \\\hline
8 & 73.71\%\ & \textbf{75.64\%}\ \\\hline
9 & 73.48\%\ & 75.41\%\ \\\hline
10 & 73.84\%\ & 75.47\%\ \\\hline
\end{tabular}
\caption{Table compares the accuracy using smaller subwindow size vs larger subwindow size.}
\label{tab:tab2}
\end{table}

As shown in Tab. \ref{tab:tab2}, method B seems to be performing better, with almost negligible increase in training time due to the higher dimension feature vector generated. Interestingly, it also seems to show there is diminishing
returns with averaging more models. Once more than 8 models are averaged
together, accuracy drops slightly. 

Next, we determine the best performing local keypoint descriptors.
We test SIFT, HSV-SIFT, Opponent-SIFT and Transformed Color SIFT.

\begin{table}[t]
\centering
\begin{tabular}{|c|c|c|c|c|}\hline
Number of models & SIFT & HSV-SIFT & Opponent-SIFT & Transformed color SIFT \\\hline
1 & 72.27\%\ & 74.22\%\ & 78.18\%\ & 79.03\%\ \\\hline
2 & 73.97\%\ & 76.64\%\ & 79.7\%\ & 81.38\%\ \\\hline
3 & 74.90\%\ & 77.44\%\ & 80.72\%\ & 82.44\%\ \\\hline
4 & 75.20\%\ & 77.89\%\ & 81.1\%\ & 82.93\%\ \\\hline
5 & 75.32\%\ & 77.91\%\ & 81.70\%\ & 82.99\%\ \\\hline
6 & 75.15\%\ & 78.01\%\ & 82.10\%\ & 82.93\%\ \\\hline
7 & 75.49\%\ & 78.20\%\ & 81.9\%\ & 82.76\%\ \\\hline
8 & 75.66\%\ & 78.88\%\ & 81.99\%\ & 82.95\%\ \\\hline
9 & 75.62\%\ & 78.86\%\ & 82.27\%\ & 83.12\%\ \\\hline
10 & 75.85\%\ & 78.94\%\ & 82.44\%\ & \textbf{83.14\%}\ \\\hline
\end{tabular}
\caption{Table compares the accuracy using different local keypoint descriptors.}
\label{tab:tab3}
\end{table}

As shown in Tab. \ref{tab:tab3}, color variants of SIFT seems to be performed significantly better than SIFT, indicating that color information is a pretty strong discriminating factor.

Finally, we investigate if combining more models would improve the accuracy
even further. In method C, we combine 10 models and set visual vocabulary size to Eq. \ref{eq:eq1}, while in method D, we combine
15 models and set visual vocabulary size to Eq. \ref{eq:eq2}. In both cases, we used Transformed Color Sift as the local keypoint 
descriptor.

\begin{equation}
\label{eq:eq1}
N=1000,1000,800,800,...,200,200
\end{equation}

\begin{equation}
\label{eq:eq2}
N=1000,1000,1000,800,800,800,...,200,200,200
\end{equation}

\begin{table}[t]
\centering
\begin{tabular}{|c|c|}\hline
Method C & Method D \\\hline
83.13\% & \textbf{83.46\%}\ \\\hline
\end{tabular}
\caption{Table compares the accuracy using different number of models.}
\label{tab:tab4}
\end{table}

As shown in Tab. \ref{tab:tab4}, perhaps unsurprisingly, combining more
models did help to improve the accuracy. We tried increasing the number more,
unfortunately, we hit a point of diminishing returns.

On a final note, the
reason having more models help obtained a higher accuracy, is most probably
due to the fact that since visual words are obtained randomly, each model's
feature was encoded with a different set of visual word, thereby giving each model a
different viewpoint of the problem.

In terms of training time, the test program which was coded in C++ and speed up with
OpenMP to take advantage of all 4 cores, took about 5 hours
to train for the 10-model classifier and about 8 hours for the 15-model classifier on
a single quad-core Intel Core i5-2500K machine.

\section{Conclusion \& Future works}
To conclude, we have presented our proposed visual object classification method that utilizes only a single best-performing local keypoint descriptor (Transformed color SIFT) while being efficient to train. We managed to obtained state of the art result on Caltech-101 of 83.46\%, showing the potential of such architecture. We suspect better results might be possible if multiple models utilizes different type local keypoint descriptor. However, it is unclear yet if significantly better results would be possible or if it would
slow down training too much.


\end{document}